\documentclass[runningheads]{llncs}

\usepackage{eccv}

\usepackage{eccvabbrv}

\usepackage{graphicx}
\usepackage{booktabs}

\usepackage{makecell}
\renewcommand\theadfont{\normalsize}
\usepackage{booktabs}
\usepackage{multirow}
\usepackage{algorithm}
\usepackage{algorithmic}
\usepackage{comment}
\usepackage{graphicx}
\usepackage{stix}

\newcommand{\gWedge}{\mathbin{\rotatebox[origin=c]{-90}{$\Wedge$}}}
\usepackage{amssymb}
\usepackage{pifont}
\newcommand{\cmark}{\ding{51}}%
\newcommand{\xmark}{\ding{55}}%

\usepackage{float}

\usepackage[accsupp]{axessibility}  % Improves PDF readability for those with disabilities.

\usepackage[pagebackref,breaklinks,colorlinks,citecolor=eccvblue]{hyperref}
\usepackage{hyperref}

\usepackage{orcidlink}

\begin{document}

\title{BAAF: Universal Transformation of One-Class Classifiers for Unsupervised Image Anomaly Detection} 

\titlerunning{BAAF}

\author{Declan McIntosh\inst{1}\orcidlink{0000-0002-7824-2339} \and
Alexandra Branzan Albu\inst{1}\orcidlink{0000-0001-8991-0999}}

\authorrunning{D.~McIntosh and A.~Albu}

\institute{University of Victoria, Victoria B.C., Canada 
\email{\{declanmcintosh,aalbu\}@uvic.ca}}

\maketitle

\begin{abstract}
    Detecting anomalies in images and video is an essential task for multiple real-world problems, including industrial inspection, computer-assisted diagnosis, and environmental monitoring. 
Anomaly detection is typically formulated as a one-class classification problem, where the training data consists solely of nominal values, leaving methods built on this assumption susceptible to training label noise. We present Bootstrap Aggregation Anomaly Filtering (BAAF), a method that transforms an arbitrary one-class classifier-based anomaly detector into a fully unsupervised method. This is achieved by leveraging the unique intrinsic properties of anomaly detection: anomalies are uncommon in the sampled data and generally heterogeneous. These properties enable us to design a modified Bootstrap Aggregation method that uses multiple independently trained instances of supervised one-class classifiers to filter the training dataset for anomalies. This transformation requires no modifications to the underlying anomaly detector; only the algorithmically selected data bags used for training change. We demonstrate empirically that our method can transform a wide variety of one-class classifier-based image anomaly detectors into unsupervised ones. Consequently, we present the first unsupervised logical anomaly detectors for images. We also demonstrate that our method achieves state-of-the-art performance in fully unsupervised anomaly detection on the MVTec AD, ViSA, and MVTec Loco AD datasets. As improvements to one-class classifiers are made, our method directly transfers those improvements to the unsupervised domain, linking the domains. 
%\textbf{Code: Link Held For Review.}
   
\end{abstract}

% Make table cells the correct size 
\renewcommand\theadfont{\footnotesize}
 
\section{Introduction}
\label{sec:intro}

% ------------- Qualitative image of results for patchcore wrapped vs not wrapped 
\begin{figure}[h]
    \centering
    \includegraphics[width=0.525\linewidth]{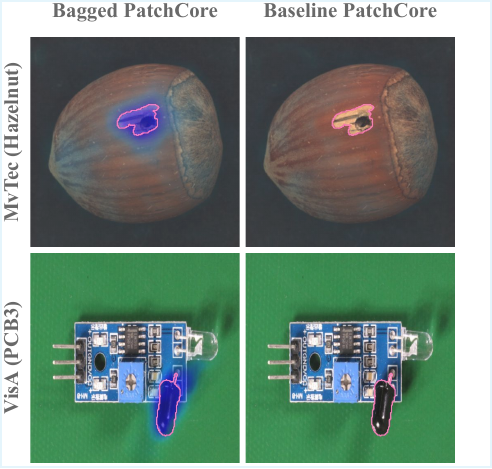}
    \caption{Comparison of prediction on training corruptions of PatchCore with and without Bootstrap Aggregation Anomaly Filtering (BAAF). Pink outlines are the anomaly ground truth, and regions of increasingly darker blue are predicted anomalies.}
    \label{fig:qualitative}
\end{figure}

% ------------------- General Field / Motivation 
Anomaly detection broadly seeks to identify rare or novel structures in data. Anomaly detection and localization for images and video is a field of key interest for many application domains, including industrial inspection, medical imaging, video surveillance, biodiversity assessment, and environmental monitoring \cite{chandola2009anomaly, bezdek2011anomaly}. In these contexts, anomalies can be defects, pathologies, or novel species behaviors; the automatic detection of each of these is highly relevant for its respective field \cite{loco_effecient, wang2024real, fernando2021deep}. However, industrial inspection remains the most studied application \cite{wang2024real, loco_effecient}.

% ------------------- OCC methods are good in general and diverse

Modern One-Class Classification (OCC)-based anomaly detection methods have demonstrated excellent performance in image anomaly detection, employing a wide range of strategies \cite{bezdek2011anomaly, nassif2021machine}. Further OCC methods have been developed to address diverse application-specific considerations, including logical anomaly detection, low-latency anomaly detection, RGB-D anomaly detection, and multi-view anomaly detection \cite{loco_effecient, bhunia2024looking, wang2024real}. These methods perform very well on their respective benchmarks but still universally use the OCC problem formulation, assuming the training dataset consists only of the nominal class \cite{loco_effecient, xisoftpatch}. While these methods can detect more diverse types of anomalies or anomalies much faster, they still require a manually curated nominal dataset; otherwise, they will learn and overfit to the present anomalies during training as nominal, leading to false negatives during inference \cite{mcintosh2023inter, xisoftpatch}. In ~\cref{fig:qualitative}, Baseline PatchCore, an OCC method, is shown missing clearly apparent anomalies because they were present during training, allowing it to overfit to these samples. This overfitting is desirable in OCC methods to create a minimal area model of nominal samples, but leads to high sensitivities to any label noise in the supervised OCC setting. This requirement for purely nominal training data is especially limiting for applications where nominal data may not be known a priori, such as in biodiversity assessment or environmental monitoring \cite{maturo2024environmental, bjerge2023hierarchical}. Furthermore, in real-world deployments, even with manually curated datasets, label noise is inevitably present \cite{long2024understanding}.

% ------------------- Existing Development of Unsupervised Anomaly Detection Methods 

Several fully unsupervised anomaly detection methods have been developed to address this, offering high tolerance to training anomalies and thereby removing the need for OCC assumptions and broadening the real-world applicability of anomaly detection methods \cite{mcintosh2023inter, im2025fun}.
However, these methods perform worse than existing OCC methods when anomalies are absent during training, resulting in a trade-off between anomaly detection performance and noise tolerance \cite{xisoftpatch, mcintosh2024unsupervised}. 
Furthermore, these methods lack the diversity of existing OCC methods, as they independently converged on patch-based approaches \cite{xisoftpatch, mcintosh2023inter}. 
Due to this convergence, adapting existing fully unsupervised anomaly detection methods to more diverse applications, such as logical anomalies, would require fundamental changes to their design \cite{im2025fun, xisoftpatch, mcintosh2023inter}.

% ------------------ Our proposed method and how it works 
To address current limitations in unsupervised anomaly detection methods, we propose Bootstrap Aggregation Anomaly Filtering (BAAF), a regularization regime for anomaly detection that transforms any existing OCC method into a fully unsupervised one, leveraging the intrinsic properties of the anomaly-detection problem. BAAF achieves this by applying a modified Bootstrap Aggregation (sometimes abbreviated as bagging) that regularizes overfitting OCC methods, which can then filter the training dataset for anomalies using intrinsic properties of the anomaly detection problem. BAAF is described in~\cref{sec:proposed}, while the necessary intrinsic properties of anomaly detection and their resulting method, motivating the correlary, are described in~\cref{sec:formulation} and~\cref{sec:corollary}. 
In this paper, our proposed transformation is applied to a diverse set of OCC methods and anomaly detection tasks, including logical anomaly detection in~\cref{sec:results}. 

% ------------------ Our methods Key Contributions
\textbf{Our work's key contributions include:}
\begin{itemize}
    \item A novel regularization, Bootstrap Aggregation Anomaly Filtering, which adapts arbitrary existing and future supervised OCC anomaly detection methods to operate fully unsupervised without modification.
    \item New state-of-the-art results in fully unsupervised anomaly detection on the MvTecAD, ViSA, and LOCO AD datasets using BAAF to augment existing OCC methods.
    \item The first fully unsupervised logical anomaly detection method in images.
    \item Universal applicability: we show that our method increases I-AUROC across 8 diverse supervised OCC methods by an average of 0.171 on MvTecAD with 10\% anomalies during training.
\end{itemize}

\section{Related Works}
\label{sec:related}

% Motivation for anomlay detection 
% ---------------- One Class Classifiers
% Goal is to show off the diversity in approaches 
% PatchCore
% Reverse Distilation++
% EfficientAD 
\subsection{One-Class Classifier Anomaly Detectors}
\label{sec:occ_related}
The vast majority of image anomaly detection methods rely on a one-class classification assumption, where all training data is assumed known from the nominal class \cite{chandola2009anomaly, bezdek2011anomaly, cui2023survey}. This allows these methods to model the entire training dataset and evaluate new data to assess its conformity with the model and detect anomalies, i.e., outliers, relative to the model \cite{roth2022towards}. There are two key components of these methods: how they model the nominal training data and how they measure the conformity of new data to this model \cite{cui2023survey}. Generally, these methods seek to tightly fit a model onto the known purely labelled nominal training data to maximize the performance by making the model's margin between nominal and anomalous data as large as possible, but this can lead to overfitting to the nominal training data and high sensitivities to training label noise. Since these components are so openly defined, a diverse literature of OCC methods exists \cite{cui2023survey}. 

Reconstruction methods aim to degrade images with synthetic anomalies, then reconstruct the original clean sample to model the nominal distribution by learning a mapping from corrupted images to their nominal counterparts \cite{zavrtanik2021reconstruction, xiang2023squid}. 
Then, at inference, they compare the unknown sample to its reconstruction, where significant differences are likely anomalies being mapped back to the learned nominal space \cite{vojir2021road, kim2020rapp, japkowicz1995novelty}. 
These methods are very similar to self-supervised methods, which also corrupt nominal training images but then learn to classify between the synthetic corruptions and the nominal images \cite{golan2018deep, hojjati2024self, zou2022spot}. 
Similarly, some methods utilize the structure of training General Adversarial Networks (GANs) to train both a model for generating convincing fake nominal images and a discriminator that distinguishes between real and synthetic images \cite{xia2022gan, liu2023anomaly}.
Then, at test time, the GAN discriminator is used to predict images as nominal or anomalous \cite{liu2023anomaly}. 

Some methods extract features from training images and apply a statistical model to predict on new images \cite{rippel2021gaussian, xie2023novel}. 
Further, some methods use an autoencoder structure, where they learn a compact latent representation of the training nominal images \cite{cheng2021improved, xiang2022hyperspectral}. Then, at test time, they pass images through the autoencoder and check for regions of poor reconstruction, which indicate that present anomalies as they were not well represented in the learned latent space \cite{cheng2021improved, loco_effecient}. 
Still, other methods employ a student-teacher architecture, where a single teacher model is trained to generate features from a large set of diverse images \cite{loco_effecient, zhang2023destseg}. Then, the student learns to mimic the teacher's outputs, but only on the nominal set of images \cite{rudolph2023asymmetric}. At test time, the difference between the student and teacher network-generated features is used to discriminate anomalies \cite{rudolph2023asymmetric, zhang2023destseg}. 
Patch-based methods use pre-trained feature extractors to describe image patches, then compare the set of nominal patch features with the test image patch features to distinguish anomalies based on high distances from training nominal features \cite{roth2022towards, yi2020patch, defard2021padim}. 

We reviewed here just a subset of the most popular and effective OCC method types for anomaly detection and localization, excluding less common methods such as normalized flow-based methods, attention-based methods, and density estimation \cite{gudovskiy2022cflow, tang2025anomaly, liu2022unsupervised}. 

% mention motivation of unsupervised anomaly detection, distinction from OCC anomaly detection 

%%%%% transition to unsupervised methods that are far less diverse   
\subsection{Unsupervised Anomaly Detectors}
Notably, all methods designed for OCC described in ~\cref{sec:occ_related} are not intended to tolerate anomalies corrupting the training dataset \cite{xisoftpatch, mcintosh2024unsupervised}.
This means that if corruption exists in the training dataset, the trained OCC method will incorporate that anomaly into its nominal model and will likely be blind to similar anomalies when deployed.
To address this, several methods have been developed to operate unsupervised, with sufficient tolerance to anomalies that they do not require guarantees of purely nominal datasets \cite{xisoftpatch, mcintosh2024unsupervised}. These methods, however, exhibit much less diversity in approach; specifically, they all operate as patch-based methods that use a memory bank of previously seen features to distinguish anomalies from nominal data \cite{mcintosh2023inter, im2025fun}. In fact, all of these methods leverage statistical relationships between nominal and anomalous patch features to either filter or downweight anomalies in the patch memory bank \cite{xisoftpatch, im2025fun}. 

% ---------------- InReach/Online-InReaCh
% Patch-based, uses patch matching accross images to find nominal features that span many images to generate nominal model 
% Updated for online operation 
InReaCh and Online-InReaCh work by building their nominal patch set from only symmetrically minimum-distance pairs of patches between images \cite{mcintosh2024unsupervised, mcintosh2023inter}. This approach assumes that nominal patches are common across images and that they should be more homogeneous than anomalies across image realizations \cite{mcintosh2024unsupervised, mcintosh2023inter}.
SoftPatch operates by performing patch-level noise discrimination, predicting the outlier score for each patch based on the Local Outlier Factor calculated from a memory bank of all patches \cite{xisoftpatch}. They then remove the top $\tau$ percentage of patches based on their outlier scores. 
Finally, FUN-AD considers that the distribution of distances between nominal-nominal patch pairs is distinct from that of anomaly-anomaly and nominal-anomaly pairs \cite{im2025fun}. Using this, they can iteratively reconstruct their memory bank of predicted nominal patches to generate a final nominal memory bank for comparisons at test time \cite{im2025fun}. 

Our proposed approach, described in ~\cref{sec:proposed} addresses the apparent lack of diversity in unsupervised anomaly detection methods by adapting any existing or future OCC method to be highly tolerant to training anomaly corruptions.  

% ---------------- SoftPatch
% patch-based, uses soft outlier re-weighting to determine hard anomalous features in training data 

% ---------------- FUN-AD
% Also patch based 
% uses the assumption that pairs of nominal-nominal and to a lesser extent nominal-anomaly features will have smaller distances on average than nominal-anomaly 

\section{Proposed Approach}
\label{sec:proposed}

% ------------- Figure showing the bagging of datasets 

\begin{figure*}[!h]
\begin{center}
\includegraphics[width=1.0\linewidth]{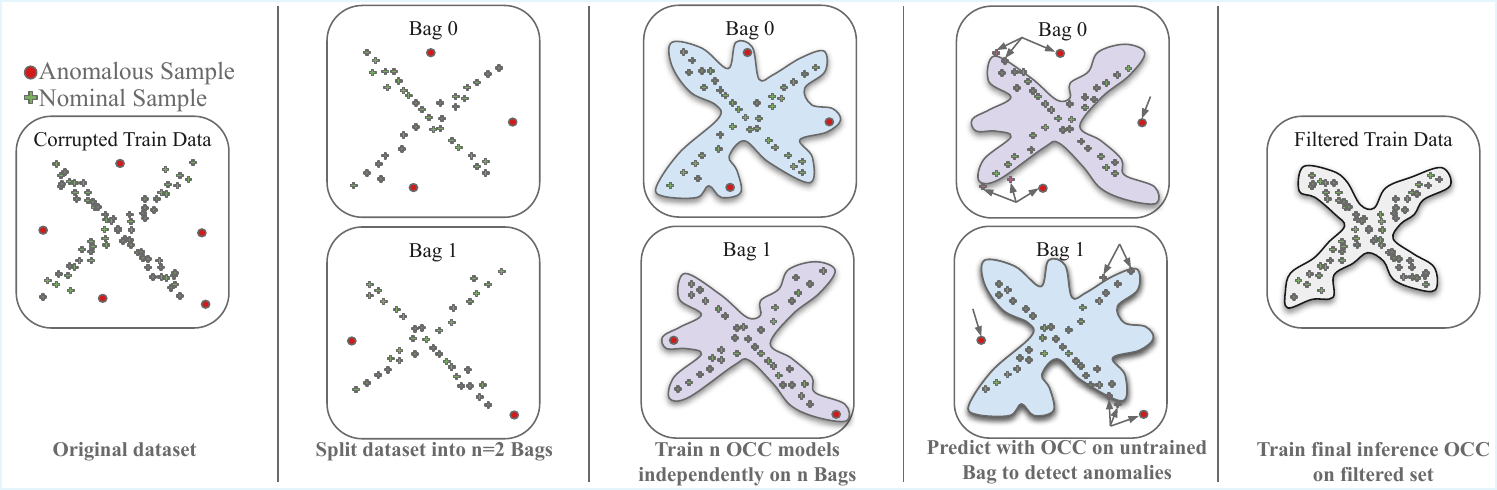}
\end{center}
   \caption{BAAF process for removing anomalies from the training dataset for unsupervised anomaly detection with $n=2$ bags. No assumptions are made about how the OCC anomaly detection model operates or the data type; our method only changes the subset of training data available to the OCC model at each step. The value $n=2$ is chosen for ease of visualization.}
\label{fig:overall}
\end{figure*}

% ---------------- Key Assumptions (The bulk ofthe  proposed approach)

BAAF transforms an arbitrary OCC supervised anomaly detection method to fully unsupervised anomaly-detection settings, where the training data may be corrupted by anomalies, thereby violating a necessary OCC assumption. This is done by leveraging fundamental properties of the anomaly detection problem presented in ~\cref{sec:formulation}, which directly motivate our corollary in ~\cref{sec:corollary}. Building from this corollary, in ~\cref{folding_procedure}, we present Bootstrap Aggregation Anomaly Filtering, BAAF, an anomaly-detection-specific extension of a modified Bootstrap Aggregation. We also present a variant of our method for video anomaly detection in ~\cref{sec:video} of the supplementary materials. An overview of the proposed BAAF method is presented in ~\cref{fig:overall}.

\subsection{Anomaly Detection Framework}
\label{sec:formulation}
Anomaly detection is applied to stochastic processes, including but not limited to the image and video domains, with some common frameworks. First, it is common to assume a uniform distribution outside the region of the normal data, no specific anomalies are more common than others \cite{MARKOU20032481, https://doi.org/10.1002/widm.1280}. Furthermore, it is common to assume that nominal data are sampled from a multivariate Gaussian distribution \cite{https://doi.org/10.1002/widm.1280, rippel2021gaussian, guo2018multidimensional}. While not perfect approximations of image distributions, these models can help motivate the general anomaly detection framework. 

\begin{equation}
|A| \gWedge |N|
\label{eq:assume_1}
\end{equation}

For Equation \ref{eq:assume_1}, $A$ is the set of all possible samples considered anomalous, $N$ is the set of all possible samples considered nominal for a given task. $N$ is a small, constrained manifold and $A$ exists as the complement of that manifold, where in the high-dimensional space of images, it is exponentially larger. This can be justified by noting that, for any nominal sample $x_n$, many possible transformations $f$ exist that corrupt it into a unique anomalous sample $x_a=f(x_n)$. For a concrete example, given any nominal image of a specific screw, there is a large set of possible unique scratches or objects placed on the screw which would constitute anomalies. From this, the set of all possible nominal images is a small subset of all possible images and, consequently, of anomalies. 

\begin{equation}
P(x_i \in N) \gWedge P(x_i \in A) | x_i\overset{\text{i.i.d.}}{\sim} F
\label{eq:assume_2}
\end{equation}

In Equation \ref{eq:assume_2}, we note that the probability of sampling an anomaly from the underlying image distribution $F$ is much lower than the probability of sampling a nominal image, which we take directly from the anomaly detection problem definition \cite{chandola2009anomaly, cui2023survey, hodge2004survey}. Anomalies are much rarer than nominal data. Notably, we assume in our formulation of ~\cref{eq:assume_2} that each sample from the image distribution $F$ is independent and identically distributed \cite{MARKOU20032481, https://doi.org/10.1002/widm.1280}. This could be violated if, for instance, a manufacturing defect introduces a consistent anomaly in all or most samples during the collection of training data. We will, however, demonstrate empirically in our testing that this independence is a weak requirement and can be relaxed. 

\subsection{Corollary}
\label{sec:corollary}

\begin{equation}
    0 \approx  P(\{x_i,x_j\} \in A) P(|x_i-x_j| < \epsilon) | x_i,x_j \overset{\text{i.i.d.}}{\sim} F
    \label{eq:conclusion}
\end{equation}

The Corollary in ~\cref{eq:conclusion} states that the probability of a pair of independent samples being both very similar to each other and both anomalous is near zero. This results directly from ~\cref{eq:assume_1} and ~\cref{eq:assume_2}.  

The first term of ~\cref{eq:conclusion} being a low value, follows directly from ~\cref{eq:assume_2}, where if the samples are independent, then $P(\{x_i,x_j\} \in A) = [P(x_i\in A)]^{2} $ when $P(x_i\in A)$ is already known low from the anomaly detection problem formulation. 

For the second term of ~\cref{eq:conclusion}, based on ~\cref{eq:assume_1}, the set of all possible anomalies is vast compared to the set of nominal images; therefore, the probability of two independently sampled anomalies being less than $\epsilon$ apart is low. In our method, $\epsilon$ denotes the given OCC anomaly detector's margin. We define margin as the minimum distance between two samples, such that if one were present during training, the other would be considered nominal by the trained OCC anomaly detector at inference.

The corollary is supported by empirical results from previous papers, in which OCC methods perform nearly indistinguishably from unsupervised anomaly detectors when the training corruptions are disjoint from the test set, as previously referred to as the 'no-overlap' case \cite{xisoftpatch}. This is because the probability that two anomalies, separated into the training and testing sets, are less than $\epsilon$ apart in the small, manually curated, and manually corrupted testing sets used is very low \cite{xisoftpatch, mcintosh2023inter}. However, for longer-term deployment, the likelihood that anomalies sampled during inference lie within $\epsilon$ of any training anomaly corruption increases substantially. Necessitating unsupervised anomaly detection methods to prevent false negatives from OCC methods overfitting to training anomalies. 

The practical result of ~\cref{eq:conclusion} is that when a limited dataset of training samples is split, there is a low chance that any pairs of anomalies exist in separate subsets of that dataset which are also less than $\epsilon$ apart. Therefore, training on a subset with one or more anomalies will likely not cause the model to predict another anomaly as nominal when predicting on a separate independent subset. This result can be used to train a model on one subset of data to filter another subset, a process formalized in ~\cref{folding_procedure}.

\subsection{Bootstrap Aggregation Anomaly Filtering}
\label{folding_procedure}

% TODO 
%These assumptions do not hold perfectly in general image anomaly detection, as anomalous images are sampled from real-world distributions in which the probability of observing a damaged item on a manufacturing line with a scratch is much higher than that of an image in which pixel values are sampled from a uniform distribution. Despite this, existing OCC image anomaly detection methods successfully project this image space onto a more manageable $\mathbb{R}^{1}$, generally in the range $[0-1]$, where these simplistic assumptions hold. This can be shown in the high AUROC scores achieved by these methods on per-image predictions in the range of $[0-1]$. In the following, we will use these properties of anomaly detection to build a regularization method to allow OCC methods to operate fully unsupervised.

Our proposed method is a modification and extension of Bootstrap Aggregation (bagging), a form of regularization in which multiple learners are trained on separate subsets of the original dataset and then vote on predictions at test time \cite{breiman1996bagging}. This method is commonly used to prevent overfitting to data, which motivates our use of it to prevent OCC methods from overfitting to anomalous data. We first modify bagging by sampling bags without replacement. This prevents a single anomaly from being sampled into multiple bags, which would violate our weak I.I.D. assumption in \cref{eq:conclusion}. We also do not use our combination of learners to make predictions, as this would greatly increase inference latency; instead, we aggregate the trained models' votes to filter the entire dataset of anomalies, then train a final OCC classifier on the resulting dataset. 
OCC methods perform well at projecting higher-dimensional images of both anomalies and nominal data onto $\mathbb{R}^{1}$ in the range $[0,1]$, removing the dimensionality challenges of images. 
In this projected space, we can assume Gaussian distributions for both anomalous and nominal samples, as it behaves very similarly to the general formulation of anomaly detection we described in~\cref{sec:formulation}. We use this modelling to generate a prediction threshold for this filtering step, defined as the crossover probability between the fitted Gaussians. Our proposed modification and extension of Bootstrap Aggregation is able to filter anomalies in this manner only by leveraging the specific properties of the anomaly-detection problem and OCC anomaly-detection methods. A full outline of our algorithm is provided in ~\cref{alg:unsupervised-filtering}.

\begin{figure}
    \centering
    \includegraphics[width=0.55\linewidth]{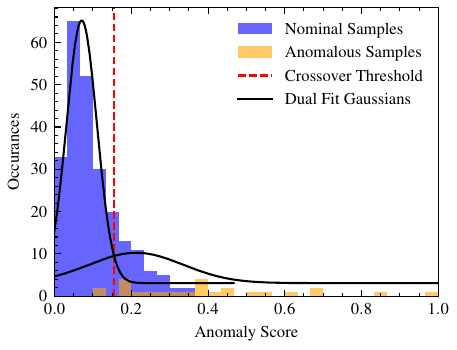}
    \caption{Example mixture of Gaussians over the predictions by PatchCore on the screw class of MvTecAD for a given bag. Notably, the assumed curve for a Gaussian distribution of nominal samples and an approximately uniform distribution of anomalies is observed.}
    \label{fig:gaussian_fits}
\end{figure}

Motivated by ~\cref{eq:conclusion}, we develop a method that uses an unmodified arbitrary OCC anomaly-detection method to filter a dataset of potentially corrupted samples. We achieve this by first splitting the dataset randomly into $n$ bags ${f_1, f_2,...,f_n}$, which are non-overlapping subsets of the original dataset, Line~\ref{line:2} of ~\cref{alg1}. The union of these bags re-forms the original dataset. We then fit a separate instance of the anomaly detection model to each bag of the original dataset (Line~\ref{line:3}). Each of these models predicts on all data, excluding the data on which it was trained (Line~\ref{line:4}). Predictions are normalized to the range $[0,1]$, where $1$ represents an anomaly and $0$ represents a nominal value (Line~\ref{line:5}). For each of these sets of predictions excluding a bag, we fit a mixture of two Gaussians (GMM) to the predictions (Line~\ref{line:6}), which are biased towards nominal predictions, as shown in ~\cref{fig:gaussian_fits} \cite{Reynolds2018GaussianMM}. We bias the GMM towards nominal features by weighting each sample by one minus its predicted anomaly value, so highly anomalous outlier samples have a lesser effect on the Gaussian fit. This inductive bias is chosen to ensure our filter has high precision, at the cost of lower recall of nominal images. Lower recall can be tolerated as OCC anomaly detectors are generally very data-efficient; therefore, the loss of nominal data is less detrimental \cite{mvtec_dataset}. The crossover point between these two Gaussian probability functions is taken as the filter threshold (Line~\ref{line:7}). This is selected such that one distribution covers nominal samples and the other distribution covers anomalous samples, and nominal samples which are not well represented in the given bag. For each training sample, we record whether it was above or below the nominal threshold. If, in the majority of these per-bag predictions, a sample was predicted to be anomalous, it is removed from the candidate training dataset (Line~\ref{line:8}). This entire process can be repeated multiple times with new random bags for k iterations, which we refer to as votes, such that the final training set is the average of the resulting datasets over multiple runs (Line~\ref{line:11}). Once a final set of images is determined to be nominal, we train a final instance of the OCC on this filtered dataset (Line~\ref{line:12}). We will denote any configuration of bags and votes in this paper as \emph{BAAF(\# of votes/\# of bags)+OCC Method.}

\begin{algorithm}
\footnotesize
\caption{\textit{BAAF($K,n$)}: Transformation of OCC method for Unsupervised Anomaly Detection}
\label{alg1}
\begin{algorithmic}[1]
\renewcommand{\algorithmicrequire}{\textbf{Input:}}
\renewcommand{\algorithmicensure}{\textbf{Output:}}
\renewcommand{\thealgorithm}{}
\REQUIRE Dataset $D$, OCC Anomaly Detection Method \emph{TrainModel}, \\ Number of bags $n$, Number of votes $K$
\ENSURE Trained OCC Anomaly Detector $M_T$

\FOR{$k \gets 1$ to $K$} \label{line:1}
    \STATE $\{D^{(i)}\}_{i=1}^n \gets \text{RandomSplit}(D, n)$ \label{line:2}
    
    \STATE $\{M^{(i)}\}_{i=1}^n \gets \{\text{TrainModel}(D^{(i)})\}_{i=1}^n$
    \label{line:3}
    
    \STATE $P^{(i,j)}[x] \gets \text{Predict}(M^{(j)}, D^{(i)}[x]), \forall\ i \in [n],\ j \in [n] \setminus \{i\},\ x \in [|D^{(i)}|]$
    \label{line:4}

    \STATE $P \gets \frac{P - \min(P)}{\max(P) - \min(P)}$
    \label{line:5}

    \STATE $G^{(i)}_{1,2} \gets \text{FitGMM}_{g=2}\Big(\{P^{(i,j)}\}_{j \neq i},  \text{weight} = \{1 - P^{(i,j)}\}_{j \neq i}\Big)$
    \label{line:6}

    \STATE $T^{(i)} \gets \text{Solve}\big(\text{PDF}(G^{(i)}_1) = \text{PDF}(G^{(i)}_2)\big)$
    \label{line:7}

    \STATE $D'^{(i)} \gets D^{(i)} \setminus \Big\{D^{(i)}[x] \mid \#\{j \in [n] \setminus \{i\} \mid P^{(i,j)}[x] > T^{(i)}\} < \frac{n-1}{2} \Big\}$
    \label{line:8}

    \STATE $D^{(k)} \gets \bigcup_{i=1}^n D'^{(i)}$
    \label{line:9}
\ENDFOR

\STATE $D_f \gets \text{MajorityVote}(\{D^{(k)}\}_{k=1}^K)$
\label{line:11}

\STATE $M_T \gets \text{TrainModel}(D_f)$
\label{line:12}

\RETURN $M_T$
\end{algorithmic}
\label{alg:unsupervised-filtering}
\end{algorithm}

This method works as the likelihood that a specific anomaly is well represented in a majority of other bags is expected to be very low. Therefore, by training our OCC anomaly detector on each bag, we are confident that despite the OCC method overfitting by design to its training data and learning any anomalies present in that bag to be nominal, these learned anomalies will not generalize to anomalies in other bags. A conceptual example of this can be seen in ~\cref{fig:overall}, where, despite each of the two models fitting entirely to anomalies present in their training bag, the anomalies in the other bag were distinct and therefore not misclassified. This method requires an OCC to be run (bags × iterations + 1) times, which significantly increases offline training time. This proposed bagging procedure makes no changes to the underlying OCC method; it only modifies the data available for each trained instance, ensuring that test-time inference remains identical. From this, any special considerations of the underlying OCC methods, such as the ability to detect logical anomalies or specific feature extractors for medical image domains, remain after the adaptation to unsupervised operation.

\section{Results}
\label{sec:results}

% ------------- Figures for the example distributions being voted on, nominal purely vs with corruptions 

\subsection{Testing Configuration}
For all our experiments, we use the proposed methods at their prescribed resolutions, but we scale the predicted masks to 256x256 and center crop them to 224x224 for reporting P-AUROC and AUPRO values \cite{loco_effecient, roth2022towards}. This is done to accommodate the lowest-resolution predictions from any method, allowing for direct comparisons across all methods on all datasets \cite{roth2022towards, xisoftpatch, mcintosh2023inter}. We make minimal modifications to specific methods to enable fully unsupervised use, such as removing hard-coded training lengths for specific target classes or the use of testing set performance to select optimal weight checkpoints at a specific epoch \cite{im2025fun, liu2023dmad}. Full details on all modifications to methods are available in the Supplemental Materials. All testing with a specific dataset and anomaly corruption proportion uses the same set of corruptions from the testing set. We conduct all our testing in the more challenging unsupervised overlapping setup, where anomaly corruptions observed during training are also present during testing \cite{loco_effecient, im2025fun}. This contrasts with the non-overlapping task, where training anomalies are absent during testing.

% Requires: \usepackage{booktabs}

\begin{table*}[!ht]
    %\centering

    \footnotesize
    \caption{Performance of unsupervised anomaly detectors on MvTecAD and ViSA datasets. BAAF(3/4)+PatchCore shows the best performance in anomaly detection (I-AUROC) on both datasets for corrupted datasets.}
    \hspace{-17pt}
    \begin{tabular}{|c|l|cccc|cccc|}
    \cline{3-10}
         \multicolumn{1}{c}{} &  & \multicolumn{4}{c|}{0\% Corruptions} & \multicolumn{4}{c|}{10\% Corruptions}  \\
    
    \cline{2-5} \cline{6-10}
           \multicolumn{1}{c|}{} & Metric   & \thead{InReaCh \\  \cite{mcintosh2023inter}}  & \thead{SoftPatch \\  \cite{xisoftpatch}} & \thead{FUN-AD \\  \cite{im2025fun}} & \thead{BAAF(3/4) \\ PatchCore \\  \cite{roth2022towards}}  & \thead{InReaCh \\  \cite{mcintosh2023inter}}  & \thead{SoftPatch \\  \cite{xisoftpatch}} & \thead{FUN-AD \\  \cite{im2025fun}} & \thead{BAAF(3/4) \\ PatchCore  \\ \cite{roth2022towards}}  \\
       \hline
       \multirow{3}{*}{\rotatebox{90}{MvTec}}   & I-AUROC  &  \emph{0.884} &	\underline{0.985} &	0.862 & \textbf{0.992} & \emph{0.855} & \textbf{0.984} & \underline{0.965} & \textbf{0.984} \\
         & P-AUROC  &  \emph{0.950} &	\underline{0.981} &	0.939 & \textbf{0.982} & 0.935 & \emph{0.960} & \textbf{0.975} & \underline{0.974} \\
         & AUPRO    &  \emph{0.825} &	\underline{0.929} &	0.748 & \textbf{0.945} & 0.805 & \underline{0.903} & \emph{0.888} & \textbf{0.932} \\
       \hline
       \multirow{3}{*}{\rotatebox{90}{ViSA}}       & I-AUROC  &  0.777 &	\underline{0.915} &	\emph{0.810} & \textbf{0.960} & 0.730 & \emph{0.893} & \underline{0.909} & \textbf{0.910} \\
              & P-AUROC  &  \emph{0.959} &	\underline{0.984} &	0.906 & \textbf{0.987} & \underline{0.944} & 0.887 & \emph{0.929} & \textbf{0.951} \\
              & AUPRO    &  \emph{0.769} &	\underline{0.901} &	0.691 & \textbf{0.939} & 0.705 & \underline{0.797} & \emph{0.757} & \textbf{0.879} \\
     \hline
     Avg. & I-AUROC & 0.831 & \underline{0.95} & \emph{0.836} & \textbf{0.976} & 0.793 & \underline{0.939} & \emph{0.937} & \textbf{0.947} \\
    \hline

    \end{tabular}

    \label{tab:vs_unsuspervised_baselines}
\end{table*}

\subsection{Results on Image Anomaly Detection}

% ------------- Unsupervised methods comparison SOTA 
% ------------- Folding diverse methods on MvTec
We present results of our method, which extends unmodified PatchCore to enable unsupervised operation, achieving SOTA results among unsupervised methods on both uncorrupted and corrupted MvTecAD and ViSA, as shown in ~\cref{tab:vs_unsuspervised_baselines}. 
We obtain results very similar to SoftPatch's on the MvTecAD dataset, as expected, since SoftPatch is a modification of PatchCore designed to extend it to unsupervised anomaly detection \cite{xisoftpatch, roth2022towards}. 
However, unlike existing baselines, we did not need to design our underlying anomaly detection method's architecture to tolerate anomalies during training \cite{mcintosh2023inter, xisoftpatch, im2025fun}. 
This enables our transformation to make not just a specific method, such as PatchCore, unsupervised, but also any existing OCC anomaly detection method.

BAAF, as shown in ~\cref{tab:diverse_working_methods}, is applicable across various methods. We employ a diverse selection of methods to demonstrate this, including memory bank-based approaches such as PatchCore and DinoAnomaly, as well as reconstruction-based methods and those that utilize student-teacher architectures, such as EfficientAD \cite{roth2022towards, guo2025dinomaly, loco_effecient}.  
Our method renders all selected existing OCC image anomaly detectors highly tolerant to training anomaly corruptions. 
Our proposed method links unsupervised anomaly detection as a complementary task to OCC anomaly detection, with improvements in OCC performance linked to improvements in the unsupervised task when augmented with BAAF, as seen in ~\cref{tab:diverse_working_methods}. 
Now, any future advancement in the highly active field of OCC anomaly detection will be a direct advancement in unsupervised anomaly detection. 

\begin{table}[h]

    \label{tab:placeholder_label}
    \caption{Performance of diverse OCC methods with and without BAAF on corrupted and uncorrupted MvTecAD dataset in image AUROC \cite{mvtec_dataset}. Supplementary materials give implementation details for each model.}
    \hspace{-15pt}
    \begin{tabular}{|l|c|c|c||c|c|c|}
        \cline{1-6}
        Corruption \% & \multicolumn{3}{c||}{10\%} & \multicolumn{2}{c|}{0\%} \\
        \cline{1-6}
        BAAF(1/4) & \cmark & \xmark & (\cmark - \xmark) & \cmark & \xmark \\
        \hline
        Metric & I-AUROC & I-AUROC & Diff & I-AUROC & I-AUROC &  $\downarrow$ OCC Method Type \\
        \hline
        %\multirow{8}{*}{\rotatebox{90}{BAAF(1/4)}}
            AnomalyDino  \cite{damm2024anomalydino}  & 0.951 & 0.671 & +0.280 & 0.986 & 0.991 & Memory Bank, Patch Based       \\
            Rev-Distil++  \cite{Tien_2023_CVPR}  & 0.957 & 0.784 & +0.173 & 0.971 & 0.985 & Distillation, Reconstruction            \\
            EfficientAD \cite{loco_effecient}   & 0.966 & 0.888 & +0.078 & 0.983 & 0.989 & Student Teacher, Autoencoder      \\
            PatchCore \cite{roth2022towards}   & 0.983 & 0.772 & +0.211 & 0.994 & 0.995 & Memory Bank, Patch Based       \\
            Dinomaly \cite{guo2025dinomaly}   & 0.987 & 0.930 & +0.056 & 0.981 & 0.992 & Reconstruction, Transformer       \\
            Dfm  \cite{ahuja2019probabilisticmodelingdeepfeatures}                   & 0.866 & 0.630 & +0.236 & 0.892 & 0.935 & Contrastive Learning              \\
            Fastflow \cite{yu2021fastflow}   & 0.899 & 0.773 & +0.126 & 0.890 & 0.915 & Normalizing Flow                  \\
            Padim  \cite{defard2021padim}  & 0.853 & 0.648 & +0.205 & 0.862 & 0.897 & Mixture of Gaussians              \\
        \hline
            Average                & 0.933	& 0.762 &	0.171 &	0.945 &	0.962 &  Various          \\
        \hline
    \end{tabular}

    \label{tab:diverse_working_methods}
\end{table}

We observe considerable improvements over the OCC baselines with 10\% anomaly corruptions in the training data, as shown in ~\cref{tab:diverse_working_methods}. These differences are most pronounced in hard decision boundary methods, such as those that use patch-based nearest neighbor comparisons, and least pronounced in methods that have fuzzier boundaries, such as student-teacher and autoencoder methods \cite{roth2022towards, loco_effecient}. Notably, the improvement in the unsupervised case does not come at a significant cost in the OCC case, unlike that observed with the previous unsupervised methods. The reductions in OCC performance in our proposed BAAF method stem from removing some nominal data from the training set, thereby reducing the size of the final set. We expect that in real-world deployments of our method, the smaller filtered set can be compensated for by collecting more data without requiring manual inspection of the training data for anomalies.

\subsection{Extension to Logical Anomalies}

% ------------- LOCO AD SOTA 
Notably, the method we selected to demonstrate the effectiveness of our BAAF procedure for student-teacher and autoencoder networks, EfficientAD, was designed for both logical and structural anomaly detection \cite{loco_effecient}. 
While structural anomalies may include foreign objects or dents, logical anomalies encompass issues such as excessive quantities of a nominal object or anomalous relationships between otherwise nominal objects. 
We assume no prior knowledge of the types of anomalies in our formulation of our BAAF method, only that the underlying OCC method performs well on the target anomalies. 
In ~\cref{tab:loco_unsuspervised} we show that our BAAF procedure can generalize to models designed for logical anomaly detection. We expect this generalization to hold for any other type of anomaly formulation.

\begin{table*}[!hb]
    \footnotesize
    \centering
    \caption{Performance of OCC supervised methods and BAAF wrapped PUAD on corrupted and uncorrupted LOCO AD dataset \cite{loco_effecient}.}
    \begin{tabular}{|c|l|c|c|c|cc|}
        \hline
        Corruption \%& Metric & SoftPatch \cite{xisoftpatch}
        & FUN-AD  \cite{im2025fun}
        & Efficient-AD  \cite{loco_effecient}
        & \multicolumn{2}{c|}{PUAD \cite{sugawara2024puad}} \\
        & & & &  & Base& BAAF(1/4) \\  
        \hline
        10\% & I-AUROC & 0.674 & 0.618 & \emph{0.790} & \underline{0.811} & \textbf{0.849}  \\
        10\% & P-AUROC & \emph{0.670} & \underline{0.688} & \emph{0.670} & 0.668 & \textbf{0.712}  \\
        10\% & AUPRO   & 0.479 & 0.392 & \underline{0.629} & \emph{0.622} & \textbf{0.647}  \\

        \hline
        0\% & I-AUROC & 0.708 & 0.530 & \emph{0.901} & \textbf{0.923} & \underline{0.911}  \\
        0\% & P-AUROC & 0.704 & 0.693 & \textbf{0.741} & \underline{0.724} & \emph{0.704}   \\
        0\% & AUPRO   & 0.501 & 0.356 & \textbf{0.681} & \underline{0.655} & \emph{0.635} \\

        \hline
    \end{tabular}

    \label{tab:loco_unsuspervised}
\end{table*}

We choose to showcase our BAAF procedure on PUAD, another OCC logical anomaly detection method, as it outperforms EfficientAD for logical anomaly detection \cite{sugawara2024puad, loco_effecient}.
BAAF(1/4)+PUAD in the 10\% anomaly corruption task shows greater performance than existing unsupervised anomaly detection baselines and considerable improvements over logical anomaly OCC baselines, see ~\cref{tab:loco_unsuspervised}. 
This is because these unsupervised baselines were not designed to handle logical anomalies and therefore cannot generalize to them \cite{loco_effecient, im2025fun, xisoftpatch}. 
While the existing logical OCC method baselines struggle with the training anomaly corruptions, which they were not designed to tolerate, degrading their test performance. The best performance in unsupervised logical anomaly detection is achieved by combining a strong logical anomaly detection method with BAAF to increase its anomaly-tolerance.

We observe that the gains in performance over the existing OCC logical anomaly detection baselines in the corrupted task are lower than in other tasks. 
This is because our BAAF procedure assumes the underlying methods are good OCC methods; however, as can be seen in ~\cref{tab:loco_unsuspervised}, the logical anomaly detection task in the uncorrupted case is more difficult than the structural anomaly detection of MVTecAD and ViSA \cite{visa_data, mvtec_dataset, loco_effecient}. 
Despite this, as better OCC methods for logical anomaly detection are released, we expect those methods, when bagged, to improve upon the unsupervised results shown today by adapting PUAD.

% ------------- Performance on range of anomaly prevelence amounts 

% Hyperparameter search for MvTecAD

\subsection{Hyperparameters}

We evaluate the selection of BAAF hyperparameters, the number of bags, and the number of votes on MvTecAD with 10\% training anomaly corruptions in ~\cref{tab:folding_hyperparameter_search}. We note a tradeoff in the number of bags used. If only two bags are used, the anomalies are not sufficiently filtered, as they are too homogeneous with each other on the MvTecAD dataset. However, if too many bags are used, there is less data to train the underlying PatchCore OCC, leading to somewhat poorer anomaly filtering. The proportion of the original dataset in each bag is $1/n$, where $n$ is the number of bags, so for the toothbrush class of MvTecAD with 8 bags, only 7 images are used to train each instance of the underlying OCC to filter anomalies \cite{mvtec_dataset}. 

\begin{table}[!h]
    \footnotesize
    \centering
    \caption{Hyperparameter search for BAAF(votes/bags)+PatchCore on MvTecAD with 10\% training corruptions using I-AUROC as the metric. \cite{roth2022towards, mvtec_dataset}}
    \begin{tabular}{|c|cccc|}
        \hline
        Votes $\downarrow$ Bags $\rightarrow$ & 2 & 4 & 6 & 8 \\
        \hline
        1 & 0.883 & \underline{0.983} & \underline{0.983} & 0.980 \\
        3 & 0.871 & \textbf{0.984} & 0.981 & 0.981 \\
        5 & 0.879 & 0.981 & \emph{0.982} & 0.980 \\
        \hline
    \end{tabular}

    \label{tab:folding_hyperparameter_search}
\end{table}

Generally, as OCC methods are intentionally designed to be prone to overfitting and nominal data are homogeneous, most methods are fairly data-efficient, so this down-sampling of the dataset size used during filtering is not a major concern. Further, the final trained model used for inference is trained on the entire dataset, excluding filtered anomalies and low-confidence nominal data.   
Regarding the number of votes, we note that, in general, using three votes rather than a single vote slightly improves performance but significantly increases training time by a factor of almost three. Increasing the number of votes to 5 does not improve scores, despite the additional training time.

\subsection{Training and Inference Speed}

One significant benefit of our method is that it makes no modifications to the underlying OCC method at inference time. This makes our method as fast as the underlying OCC method used. In fact, for methods which use a memory bank, such as PatchCore or DinoAnomaly, our BAAF procedure provides a slight increase in inference speed of around 2\%. Specific results are shown in the supplemental materials. \cite{roth2022towards, guo2025dinomaly}. This is because we remove anomalies and poorly reinforced nominal data, making the memory bank smaller at inference time and reducing latency. 

This parity with existing OCC methods at inference time does not hold for our method during training. The BAAF procedure requires retraining the model multiple times to predict anomalies in other bags. Generally, the training time is $t(k\times n+1)$ where $t$ is the time to train the model once, $k$ is the number of votes, and $n$ is the number of bags. The additional 1 is the time to train the final model for inference on the filtered dataset, which takes $t\times k\times n$ to be filtered. For our main configuration of 1 vote and 4 bags, this means we need to re-train the underlying OCC 5 times. 

% Tolerance to different noise rates 

\subsection{Tolerance to Training Anomaly Rates and Sample Dependence}

In ~\cref{tab:corruption_metrics}, we evaluate our BAAF procedure on the MvTecAD dataset with varying anomaly training corruption proportions using PatchCore with 3 votes and 6 bags. Our method exhibits strong unsupervised performance, even in the worst-case scenario, where 40\% of the training images contain anomalies. We test up to 40\% anomalies, beyond which most classes are saturated, since we have included all provided anomalies as train corruptions. We see a steady decrease in performance as the proportion of anomalies in the training data increases. This is expected, as it begins to violate some of the assumptions underlying BAAF, namely that anomalies are very infrequent in the dataset. 
The difficulty at high anomaly corruption rates is exacerbated by the fact that all anomalies in the MvTecAD dataset used for this analysis are manually constructed to fit into a small set of anomaly categories; for example, the hazelnut class has only four anomaly categories \cite{mvtec_dataset}. This construction of anomalies to fit discrete categories violates another assumption of BAAF: that the images being trained on are sampled independently. 

\begin{table}[!h]
    \centering
    \caption{Performance under different anomaly corruption percentages on MvTecAD with BAAF(3/6) PatchCore, 6 bags were used to increase tolerance to very high rates of anomalies \cite{mvtec_dataset,roth2022towards}. *Performance with sampled anomalies all from the same MvTecAD category, violating the I.I.D. assumption, was evaluated using BAAF(1/4)+PatchCore.}
    \footnotesize
    \begin{tabular}{|l|ccccc||c|}
        \hline
        Corruption \% & 0.00 & 0.10 & 0.20 & 0.30 & 0.40 & Not I.I.D.* (12\%)\\
        \hline
        I-AUROC & 0.992 & 0.981 & 0.972 & 0.949 & 0.938 & 0.978 \\
        P-AUROC & 0.982 & 0.972 & 0.953 & 0.940 & 0.925 & 0.972 \\
        AUPRO   & 0.946 & 0.929 & 0.919 & 0.901 & 0.885 & 0.931 \\
        \hline
        Filter Precision & $\emptyset$ & 0.756 & 0.842 & 0.878 & 0.891 & 0.694\\
        Filter Recall    & $\emptyset$ & 0.935 & 0.944 & 0.925 & 0.921 & 0.938\\
        \hline
    \end{tabular}

    \label{tab:corruption_metrics}
    
\end{table}

We explicitly test the worst-case scenario that violates our I.I.D. assumption on MVTecAD by corrupting the test set with the largest-anomaly category for each class and including all anomalies categorized as "combined" where multiple anomaly types are present \cite{mvtec_dataset}. When this is performed, we observe an average of 12\% anomalies in the training data and present the results for BAFF(1/4)+PatchCore under these conditions in ~\cref{tab:corruption_metrics}. Under this regime, our model performs worse than in the randomly sampled 20\% anomalies case, indicating that the violation of the I.I.D. assumption reduces our regularization performance. Despite this reduction in relative performance, even categorically similar anomalies remain distinct in the high-dimensional image space, yielding strong absolute performance with an I-AUROC of 0.978 in this "Not I.I.D" case. 
Despite these challenges in the dataset, BAAF consistently maintains a high anomaly recall of over 92\% removed across all tested corruption rates and assumption violations. This indicates that these assumptions are relatively soft requirements for our method to perform well, likely due to the high dimensionality of images.

\section{Future Work and Limitations}

Our method requires significant increases in training time to supplement existing OCC methods with high tolerance to anomalies, on the order of 5-13 times longer across all of our testing configurations. This increase in training time is prohibitive for some domains, which may require online fine-tuning of the nominal model. Online unsupervised anomaly detection would be an interesting avenue for future work. Furthermore, since we make no changes to the underlying OCC methods, any limitations of the underlying method also apply to the method after augmentation with BAAF. For instance, some of the OCC methods used in this paper to achieve our state-of-the-art results rely on large pre-trained feature extractors, which limit their applicability to domains with image distributions similar to those of the pre-training tasks \cite{roth2022towards, guo2025dinomaly}. 

\section{Conclusion}
\label{sec:conclusion}
% limitations of methods based on their use of pre-trained feature extractors, but generally low level features are used as in patchcore
% 

In this paper, we introduce Bootstrap Aggregation Anomaly Filtering, which adapts arbitrary OCC anomaly detection methods to operate in a fully unsupervised manner, thereby retroactively removing the one-class assumptions of existing anomaly detection methods. Furthermore, as future improved OCC methods are developed, they will also be able to operate fully unsupervised using this procedure. Any improvement in the OCC state-of-the-art is likely to correlate with a direct improvement in the unsupervised state-of-the-art presented here. Finally, this transformation has also expanded the set of possible applications of fully unsupervised anomaly detection methods to the full scope of OCC methods, for instance, presenting the first fully unsupervised logical anomaly detection method.

\newpage

\bibliographystyle{splncs04}
\bibliography{main}

% Supplemental.
\clearpage
\setcounter{page}{1}
\section{Qualitative Comparisons}

\begin{figure*}[!h]
\begin{center}
\includegraphics[width=1.0\linewidth]{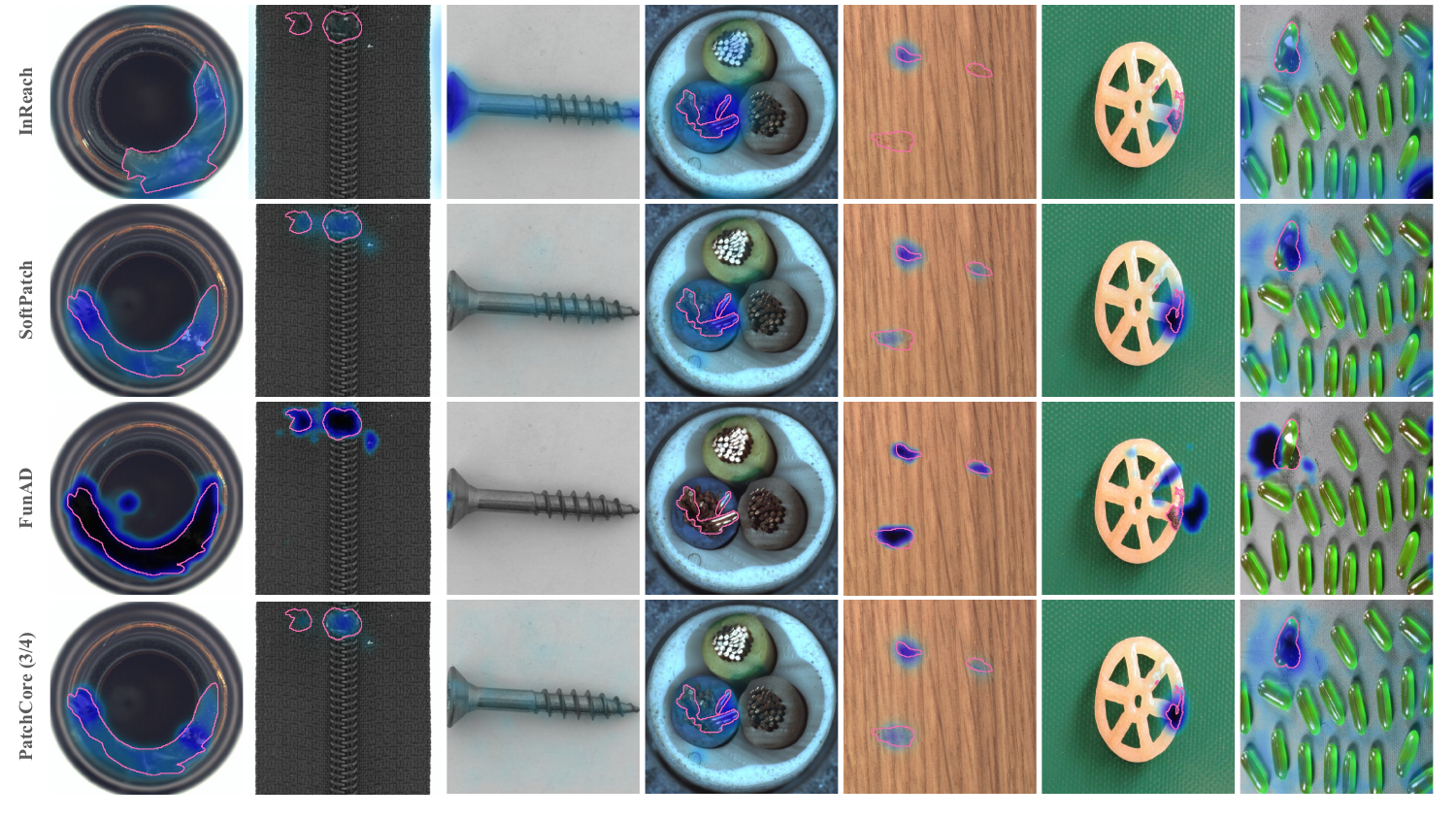}
\end{center}
   \caption{Qualitative results of unsupervised methods on MvTecAD and VisA\cite{visa_data}.}
\label{fig:qual_struc}
\end{figure*}

\begin{figure*}[!h]
\begin{center}
\includegraphics[width=1.0\linewidth]{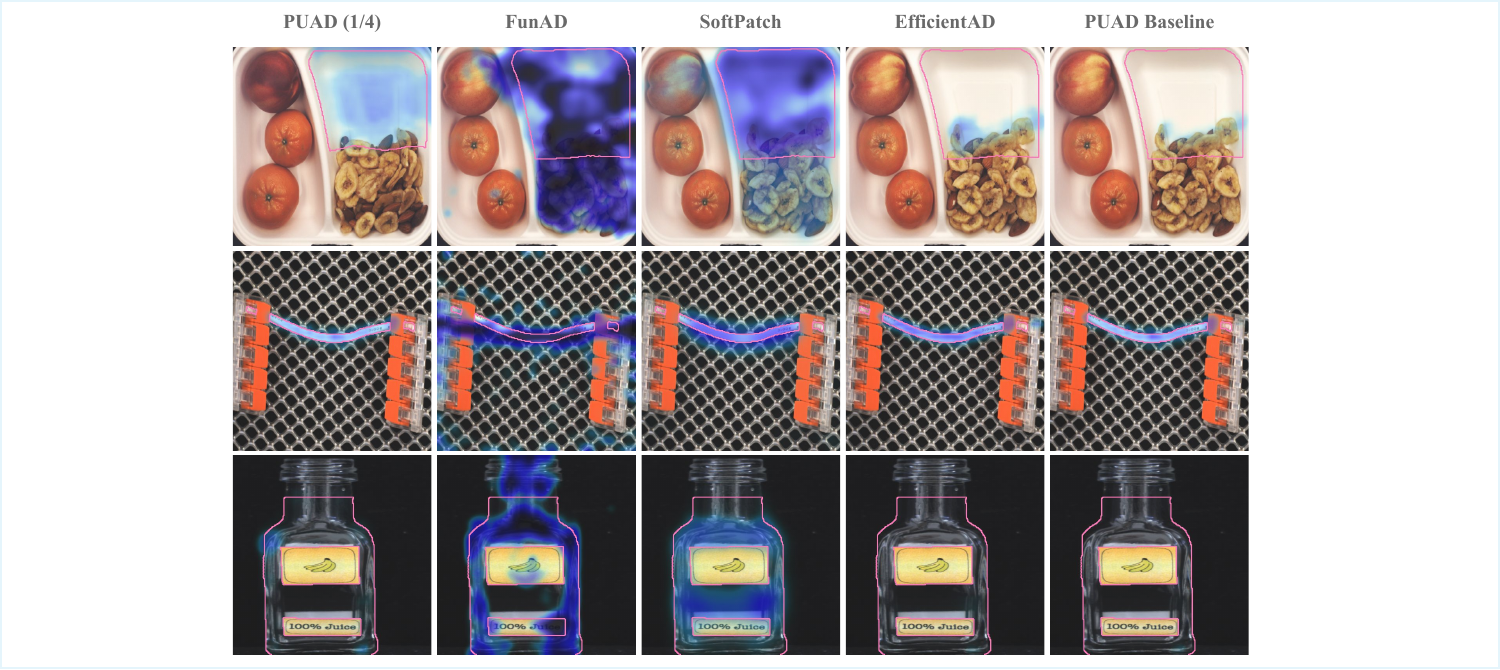}
\end{center}
   \caption{Qualitative results of unsupervised baselines (SoftPatch, FUN-AD), and OCC logical baselines (PUAD, EfficientAD) on Loco AD dataset \cite{loco_effecient}.}
\label{fig:qual_loco}
\end{figure*}

We show some additional qualitative comparisons between our method and other baselines. Images and plots are best viewed zoomed in. Images were selected to maximize the difference in anomaly predictions, thereby making the selected images more informative. See Figures~\ref{fig:qual_struc}~\&~\ref{fig:qual_loco}.

As expected from the quantitative results and the similar underlying architectures, both BAAF(1/4)+PatchCore and BAAF(1/4)+SoftPatch perform very similarly despite filtering anomalies at very different stages, image-wise against patch-wise. Both FUN-AD and InReaCh show worse qualitative performance, with occasional false negatives and false positives in the selection, though selecting for differences in prediction overstates the relative difference in actual performance.

In the first row of the logical case, showing breakfast boxes, in ~\cref{fig:qual_loco}, it can be seen that the unsupervised methods not tuned for logical anomaly detection greatly overestimate the anomalous region. In contrast, the OCC logical methods greatly underestimate it. In the second row of splicing connectors, you can see that all methods detect this structural anomaly, but the patch-based methods are generally more confident in their detection. In the bottom row of the figure showing results on a juice bottle missing its juice, this image was used during training, and both the unsupervised non-logical methods find the anomaly well, with some over-prediction. Still, the OCC logical methods miss this anomaly entirely, whereas BAAF(1/4)+PUAD shows some detection of it.

\section{Application to Video Data}

\subsection{Extension for Video Data}
% TODO modifications to the method for video data that is predicted per frame 

% only consider in our final training set clips of greater than 5 frames
% train model on all video clips, use the per-frame predictions to remove parts of clips deemed anomalous using the same procedure for images. 
% we then morphological opening then closing to remove holes in predictions, we assume that if a frame is flanked by anomlies then that frame should be an anomaly as well. 
\label{sec:video}

Generalizing our method for video requires changing the consideration of images to frames of video clips. 
To ensure independence between the bagged subsets, we split the entire video clips into separate subsets rather than individual frames. This is because each frame is highly dependent on its adjacent frames, but each video clip is independent of the others. So a single anomalous object across multiple frames in one video is not split onto multiple subsets, which we assume are independent. We then train the OCC video anomaly detection method as usual on each subset and complete the entire BAAF procedure to determine anomalous frames to remove from the training set. 

Once we have our filtered set of predicted anomalous frames, we perform a 2D morphological closing operation to reduce holes in the predicted anomalous frame regions, as we expect a frame flanked by anomalous frames also to be anomalous. Clips split by one or more regions of anomalous frames are split into separate clips in the final filtered training dataset. Finally, any clips shorter than 5 frames are discarded. Excluding these post-processing changes, all procedures for video dataset BAAF are the same as described in Algorithm~\ref{alg1}.

% ------------- Video Results 
\begin{table}[!h]
    \centering
    \begin{tabular}{l|cc}
        \hline
        \# of Corrupted Videos  & 0 & 6  \\
        \hline
        BAAF(3/2)+DMAD & 0.948 & 0.949 \\
        Baseline DMAD     & 0.944 & 0.928 \\
    \end{tabular}
    \caption{Performance of the DMAD video anomaly detector on the Peds2 dataset in frame-wise AUROC \cite{liu2023dmad, peds2_dataset}. The corrupted Peds2 dataset had 6 anomalous videos included in the training set. BAAF(3/2)+DMAD used 2 bags and 3 votes \cite{liu2023dmad, peds2_dataset}.}
    \label{tab:video_working}
\end{table}

For completeness in ~\cref{tab:video_working}, we show results for BAAF(3/2)DMAD, an OCC video anomaly detection method, on the Peds2 dataset \cite{liu2023dmad, peds2_dataset}. 
Again, our method makes no assumptions about the underlying anomaly types or data types, allowing it to be extended to video as described in ~\cref{sec:video}. 
DMAD demonstrates strong anomaly tolerance within its end-to-end fuzzy-decision-boundary deep learning architecture; however, this tolerance is further improved when operating on a corrupted dataset using BAAF.
BAAF even shows improvement in the OCC case, as we expect it to remove parts of videos that are poorly represented as nominal in the rest of the training data. 
In the case of corrupted training data, we further improve upon the OCC case by integrating additional nominal frames from the corrupted videos into the training data while pruning anomalies. 
This paper's primary focus is not on video anomaly detection, because existing OCC video anomaly detectors generally rely on pre-trained human detectors or pose estimation as an initial task for human-focused anomaly detection \cite{ramachandra2020survey, wang2022video, ahn2024videopatchcore, mishra2024skeletal}. We consider this a separate problem from the purely unsupervised general anomaly detection on which this paper seeks to improve.

\section{Ensemble BAAF}

Since each subset of our method trains an OCC method independently of the others, all the trained OCC methods need not be the same. We considered that an ensemble of models trained on each subset might perform better, similar to boosting. We found in testing that an ensemble of PatchCore \cite{roth2022towards}, DinoAnomaly \cite{guo2025dinomaly}, Reverse Distillation++ \cite{Tien_2023_CVPR}, and EfficientAD \cite{loco_effecient}, then training a final PatchCore model on the filtered dataset, performed worse than using PatchCore on its own, see ~\cref{supp:ensemble}. In the future, though, it may be best to use one slower but better anomaly detector to filter the dataset, then train a final, worse anomaly detector with faster inference for deployment.

\begin{table}[!h]
    \centering
    \begin{tabular}{l| c  c }
        \hline
        Metric & Ensemble (1/4) & PatchCore (1/4) \\
        \hline
        I-AUROC & 0.972 & \textbf{0.983} \\
        P-AUROC & 0.957 & \textbf{0.974}  \\
        AUPRO   & 0.917 & \textbf{0.935}  \\
    \end{tabular}
    \caption{Performance of Ensemble(1/4) consisting of PatchCore \cite{roth2022towards}, DinoAnomaly \cite{guo2025dinomaly}, Reverse Distillation++ \cite{Tien_2023_CVPR}, and EfficientAD \cite{loco_effecient} compared against BAAF(1/4)+PatchCore on MvTecAD \cite{mvtec_dataset} with 10\% training corruptions.}
    \label{supp:ensemble}
\end{table}

\section{Additional BAAF Prediction Distributions and Gaussian Fits}

\begin{figure*}[h!]
    \centering
    
    % ----- Row 1 (5 wide) -----
    \begin{minipage}{0.19\textwidth}
        \centering
        \includegraphics[width=\linewidth]{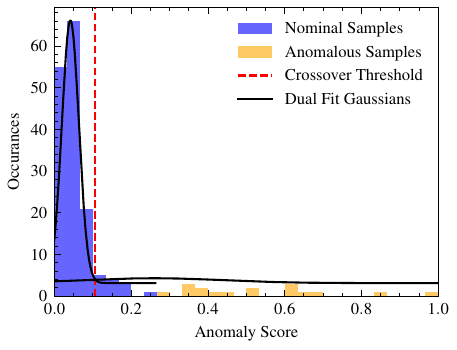}
        \caption*{Bottle}
    \end{minipage}
    \begin{minipage}{0.19\textwidth}
        \centering
        \includegraphics[width=\linewidth]{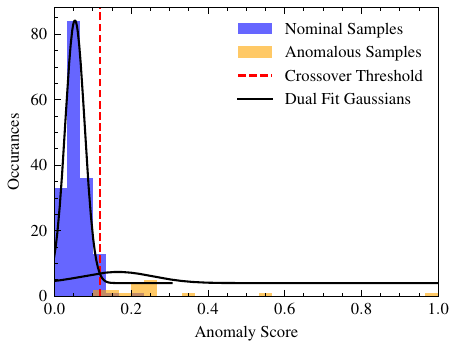}
        \caption*{Cable}
    \end{minipage}
    \begin{minipage}{0.19\textwidth}
        \centering
        \includegraphics[width=\linewidth]{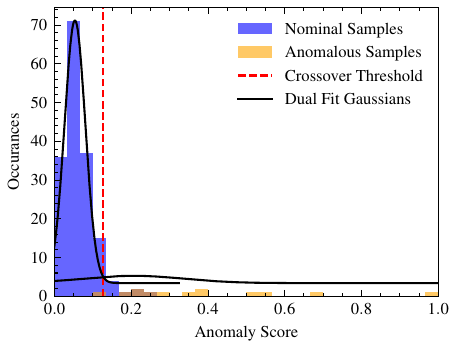}
        \caption*{Capsule}
    \end{minipage}
    \begin{minipage}{0.19\textwidth}
        \centering
        \includegraphics[width=\linewidth]{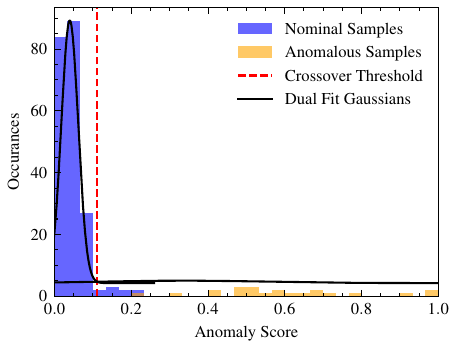}
        \caption*{Carpet}
    \end{minipage}
    \begin{minipage}{0.19\textwidth}
        \centering
        \includegraphics[width=\linewidth]{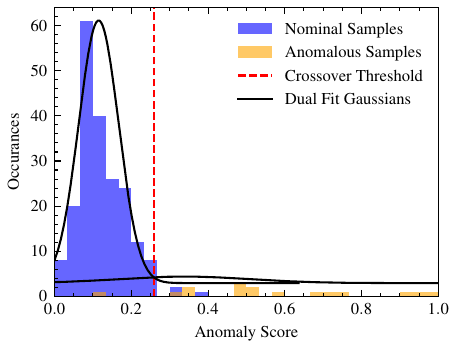}
        \caption*{Grid}
    \end{minipage}
    
    \vspace{0.3cm}
    
    % ----- Row 2 (5 wide) -----
    \begin{minipage}{0.19\textwidth}
        \centering
        \includegraphics[width=\linewidth]{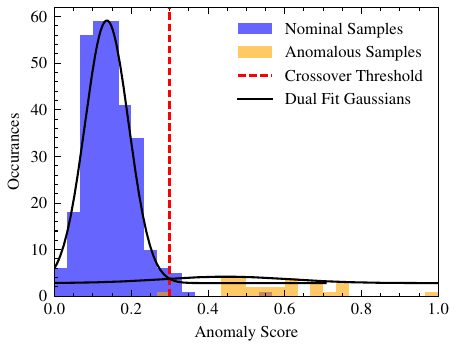}
        \caption*{Hazelnut}
    \end{minipage}
    \begin{minipage}{0.19\textwidth}
        \centering
        \includegraphics[width=\linewidth]{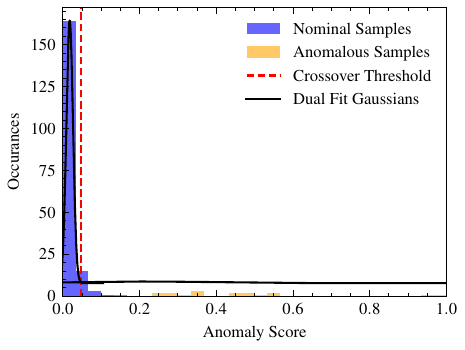}
        \caption*{Leather}
    \end{minipage}
    \begin{minipage}{0.19\textwidth}
        \centering
        \includegraphics[width=\linewidth]{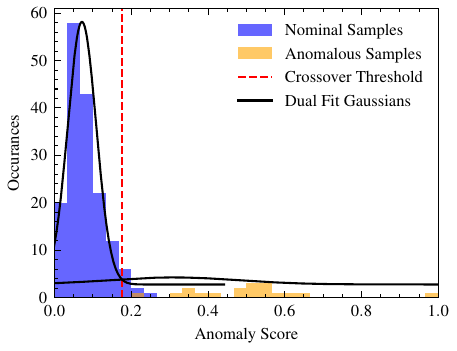}
        \caption*{Metal Nut}
    \end{minipage}
    \begin{minipage}{0.19\textwidth}
        \centering
        \includegraphics[width=\linewidth]{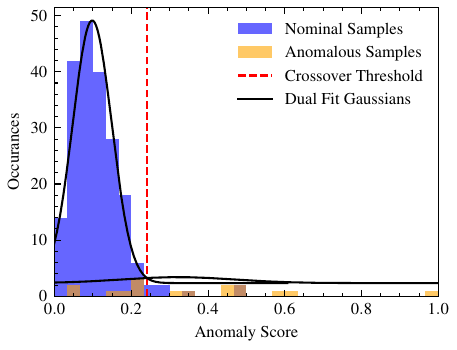}
        \caption*{Pill}
    \end{minipage}
    \begin{minipage}{0.19\textwidth}
        \centering
        \includegraphics[width=\linewidth]{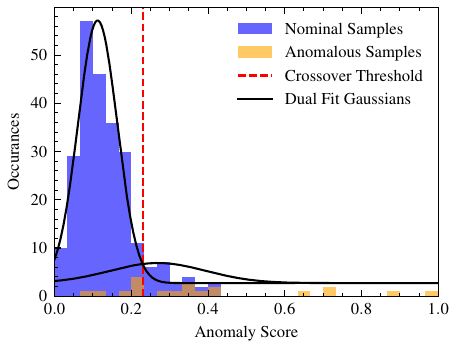}
        \caption*{Screw}
    \end{minipage}
    
    \vspace{0.3cm}
    
    % ----- Row 3 (5 wide) -----
    \begin{minipage}{0.19\textwidth}
        \centering
        \includegraphics[width=\linewidth]{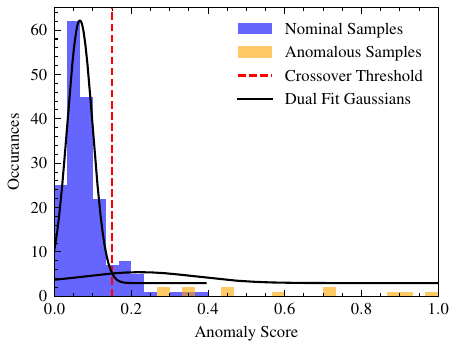}
        \caption*{Tile}
    \end{minipage}
    \begin{minipage}{0.19\textwidth}
        \centering
        \includegraphics[width=\linewidth]{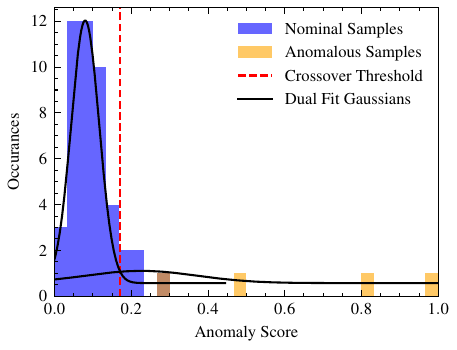}
        \caption*{Toothbrush}
    \end{minipage}
    \begin{minipage}{0.19\textwidth}
        \centering
        \includegraphics[width=\linewidth]{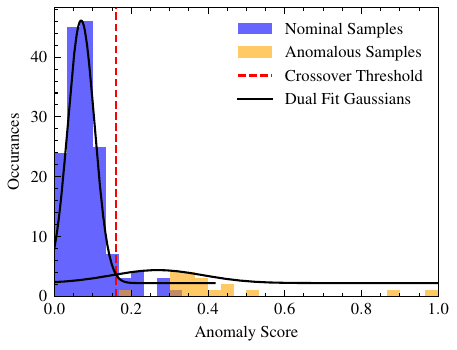}
        \caption*{Transistor}
    \end{minipage}
    \begin{minipage}{0.19\textwidth}
        \centering
        \includegraphics[width=\linewidth]{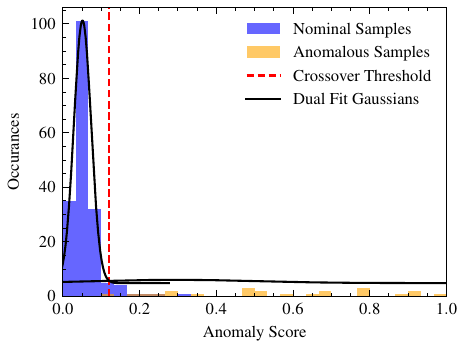}
        \caption*{Wood}
    \end{minipage}
    \begin{minipage}{0.19\textwidth}
        \centering
        \includegraphics[width=\linewidth]{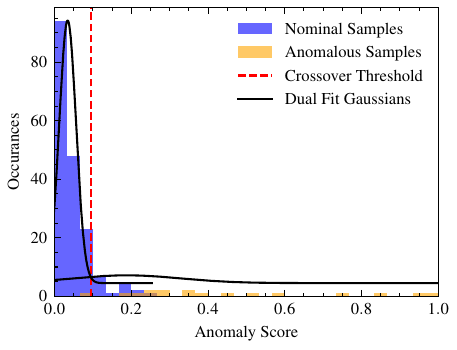}
        \caption*{Zipper}
    \end{minipage}

\caption{Example Fit Gaussian distributions on data from the first subset and the first vote of all classes in the MvTecAD dataset \cite{mvtec_dataset} with PatchCore(3/4) \cite{roth2022towards}. Determined thresholds for the anomaly filter for that subset are shown in a red dotted line. Best viewed zoomed in digitally.}
\label{supp:gaussians}
\end{figure*}

In ~\cref{supp:gaussians}, it can be observed that the large majority of anomalies present for that subset are correctly predicted as anomalies by the filter. Notably, the normalized confidence threshold differs significantly across classes, and even by manual inspection, it is clear that a fixed threshold would not perform well. This motivated the use of a Mixture of Gaussians to set the anomaly filter cutoff threshold dynamically. It is also beneficial that dynamically determining this filter threshold removes a hyperparameter from the method.  

For some classes, such as the Pill class, it can be seen that some anomalies have very low predictions, well into the distribution of the nominal sample's predictions. These samples are not filtered, as the OCC method does not pick up on the anomalies in those images. Since the model used to predict on these images was trained on a separate, independent set, even if those anomalies were not in the training data, the underlying OCC method is likely to miss them. These issues can likely only be addressed by an even better future OCC method beyond the scope of this work.  

\section{BAAF augmenting Already Fully Unsupervised Methods}

Another question we tested but determined did not work well was whether using BAAF to wrap existing unsupervised anomaly detection methods would improve their performance in the unsupervised case. To test this in ~\cref{supp:folding_alread_unsupervised}, we tested each of the existing unsupervised baselines with BAAF(1/4) and unmodified on the MvTecAD dataset with 10\% training corruptions. We found little to no improvement for both InReaCh and SoftPatch, as they both are already good anomaly filters, and adding another filter on top only reduced the apparent dataset size for these methods. FUN-AD had a much worse response to the apparent dataset reduction from the BAAF method, performing much worse when using BAAF(1/4), indicating that it is less sample efficient than the other methods for building its nominal model. 

\begin{table*}[]
    \centering
    \begin{tabular}{l| c c | c c | cc  }
        \hline
        Metric & InReaCh & InReach(1/4) & SoftPatch & SoftPatch(1/4) & FUN-AD & FUN-AD(1/4) \\
        \hline
        I-AUROC & 0.855 & 0.862 & 0.984 & 0.976 & 0.965 & 0.837 \\
        P-AUROC & 0.935 & 0.944 & 0.960 & 0.979 & 0.975 & 0.934  \\
        AUPRO   & 0.805 & 0.804 & 0.903 & 0.925 & 0.888 & 0.730  \\
    \end{tabular}
    \caption{Performance of Ensemble(1/4) consisting of PatchCore \cite{roth2022towards}, DinoAnomaly \cite{guo2025dinomaly}, Reverse Distillation++ \cite{Tien_2023_CVPR}, and EfficientAD \cite{loco_effecient} compared against BAAF(1/4)+PatchCore on MvTecAD \cite{mvtec_dataset} with 10\% training corruptions.}
    \label{supp:folding_alread_unsupervised}
\end{table*}

\section{Implementation Details of Baseline Methods}
\label{sec:implementation}

For all presented results, there are no modifications of the underlying methods when they are used with BAAF and without. 

\subsection{InReaCh \cite{mcintosh2023inter}}
We use the official code for InReaCh without modification.

\subsection{SoftPatch \cite{xisoftpatch}}
We use the official code for SoftPatch without modification.

\subsection{FUN-AD \cite{im2025fun}}
We use the official code for FUN-AD with modifications. Firstly, we removed the original implementation's use of the testing set as a validation set for selecting the best-performing checkpoint for reporting results. Since we are using this method as a baseline for existing unsupervised methods, test set labels cannot be used during training. We instead train for a number of epochs that gives approximately 40000 training steps. We determined this length based on the number of steps used to reach convergence on all classes of the MvTec AD dataset. FUN-AD also uses a synthetic dataset of corrupted nominal images from the MvTec AD dataset to report its performance on MvTec AD. Since no script was provided to generate these images for other datasets we use, such as VisA, we choose to exclude these additional training images. This leads to poor performance of FUN-AD in the 0\% corruption cases, as the method assumes anomalies exist in the training data. 

\subsection{PatchCore \cite{roth2022towards}}
We use the official code for PatchCore without modification. We choose to use the PatchCore ensemble variation for it's higher performance. 

\subsection{AnomalyDino \cite{damm2024anomalydino}}
We use the official code for AnomalyDino without modification. 

\subsection{Reverse Distillation++ \cite{Tien_2023_CVPR}}
We use the official code for Reverse Distillation++ with modifications to the training length. The original method uses a custom number of training epochs for each MvTec AD class. We fix the maximum training epochs at 300, with early stopping on a plateau of loss for 15 epochs. We find minimal change in performance with this change, but it removes the reliance on a manually set training length for each dataset. 

\subsection{EfficientAD \cite{loco_effecient}}
We use the unofficial code released on GitHub (nelson1425/EfficientAD), which closely replicates the results of the original paper. No modifications to the code's core functionality were made to adapt it to BAAF. 

\subsection{DMAD \cite{liu2023dmad}}
We use the official code, which shows very close replication to the original paper. Modifications were made to remove the code components that use test set data to select the best checkpoint during training, and instead use a fixed training duration of 10 epochs. Removing access to testing data for training broke replication of the original paper's results for image anomaly detection, so it was excluded from our comparisons. 

\subsection{PUAD \cite{sugawara2024puad}}
We use the official code for PUAD without modification.

\subsection{Dinomaly \cite{guo2025dinomaly}}
We use the Anomalib implementation of Dinomaly with their default recipe, the only modification being the removal of early stopping based on test-set evaluation. 

\subsection{Dfm \cite{ahuja2019probabilisticmodelingdeepfeatures}}
We use the Anomalib implementation of Dfm with their default recipe, the only modification being the removal of early stopping based on test-set evaluation. 

\subsection{Fastflow \cite{yu2021fastflow}}
We use the Anomalib implementation of Fastflow with their default recipe, the only modification being the removal of early stopping based on test-set evaluation. 

\subsection{Padim \cite{defard2021padim}}
We use the Anomalib implementation of Padim with their default recipe, the only modification being the removal of early stopping based on test-set evaluation. 

\end{document}